\def\year{2018}\relax
\documentclass[letterpaper]{article} 
\usepackage{aaai18}  
\usepackage{times}  
\usepackage{helvet}  
\usepackage{courier}  
\usepackage{url}  
\usepackage{graphicx}  

\usepackage{enumerate}
\usepackage{amsmath}
\DeclareMathOperator{\Tr}{Tr}
\usepackage{amssymb}
\usepackage{amsthm}
\usepackage{subcaption}
\usepackage{bbm}   

\usepackage{multirow}
\usepackage[ruled ]{algorithm2e}

\title{From Hashing to CNNs: Training Binary Weight Networks via Hashing}
\author{ Qinghao Hu,\  Peisong Wang \and Jian Cheng \\
 Institute of Automation, Chinese Academy of Sciences, Beijing, China\\
University of Chinese Academy of Sciences, Beijing, China\\
 Center for Excellence in Brain Science and Intelligence Technology, CAS, Beijing, China\\
 \{qinghao.hu,\ peisong.wang,\ jcheng\}@nlpr.ia.ac.cn
}
\begin{document}
\maketitle
\begin{abstract}
Deep convolutional neural networks (CNNs) have shown appealing performance on various computer vision tasks in recent years. This motivates people to deploy CNNs to real-world applications. However, most of state-of-art CNNs require large memory and computational resources, which hinders the deployment on mobile devices. Recent studies show that low-bit weight representation can reduce much storage and memory demand, and also can achieve efficient network inference. To achieve this goal, we propose a novel approach named BWNH to train Binary Weight Networks via Hashing. In this paper, we first reveal the strong connection between inner-product preserving hashing and binary weight networks, and show that training binary weight networks can be intrinsically regarded as a hashing problem. Based on this perspective, we propose an alternating optimization method to learn the hash codes instead of directly learning binary weights. Extensive experiments on CIFAR10, CIFAR100 and ImageNet demonstrate that our proposed BWNH outperforms current state-of-art by a large margin. 
\end{abstract}
\section{Introduction}
   Since Alexnet \cite{krizhevsky2012imagenet} made a success in ILSVRC2012 \cite{russakovsky2015imagenet}, deep convolutional neural networks have become more and more popular.  
 After that, various CNN models have been proposed such as VGGNet \cite{simonyan2014very}, Inception  \cite{szegedy2016rethinking}, ResNet \cite{he2015deep} and so on. Nowadays, these CNN models have been playing an important role in many computer vision areas 
 \cite{krizhevsky2012imagenet,ren2015faster,long2015fully}.
  
Attracted by the great performance of CNN models, many people try to deploy CNNs to real world applications. Yet the huge computational complexity and large parameter size make    CNN models hard to deploy on resource limited devices such as mobile phones and embedded devices. The huge computational complexity of CNN models makes the inference phase very slow, which is unacceptable for many real-time applications. The large parameter size brings three difficulties. First, the large parameter size means that deploying CNN models will consume huge disk storage. Second, much run-time memory is required, which is limited in many mobile devices. Third, large parameter size will cause heavy DRAM access, which consumes more energy. Since battery power is very limited in many mobile devices, this severely affects devices' battery life.

To alleviate these problems, a variety of methods have been proposed to reduce the parameter size or accelerate the inference phase. These methods can be divided into three main categories: low-rank decomposition based methods, pruning-based methods, and quantization based methods. 
Low-rank decomposition based methods \cite{denton2014exploiting,jaderberg2014speeding,zhang2015accelerating,wang2016accelerating}  decompose a weight matrix (tensor) into several small weight matrices (tensors). These methods achieve good speed-ups for large convolutional kernels, but usually perform poorly for small kernels. Besides, the compression ratio of parameters is kind of low  by using low-rank based methods. 
Network pruning has a long history and is still a widely used technique for CNN compression and acceleration \cite{han2015learning,liu2015sparse}. The main idea of these methods is to remove low-saliency parameters or small-weight connections \cite{han2015learning}. In general, after the pruning step, k-means clustering and Huffman coding are also required to make a good compression ratio. The k-means clustering and Huffman coding bring inconvenience for inference phase since we have to decode the Huffman codes and use lookup table for k-means dictionary. As a result, the decoding and lookup table will bring extra memory and computational overhead. 
Quantization based methods include codebook based quantization methods and low-bit weight representation. Codebook based quantization methods mainly use vector quantization algorithms such as K-means, Product Quantization and so on \cite{gong2014compressing,wu2016quantized} to quantize the weight kernels. These methods require lookup tables to store the dictionary, and is unfriendly to cache memory since accessing lookup tables is random and unordered.   
Low-bit weight representation methods \cite{lin2015fixed,gupta2015deep,rastegari2016xnor,dong2017learning} represent weights as low-bit fixed point even binary values. 
Low-bit weight representation can reduce run-time memory and storage demand as no decoding or lookup tables are required. As a special case of low-bit weight representation, binary weight can achieve about 32$\times$ compression ratio. In addition, since weights are represented by binary values, multiplication operations can be replaced by addition and subtraction operations. Thus binary weight can also speed up the inference phase. Nevertheless,current weight binarization methods usually will bring significant network accuracy drop, especially for large CNN models.

In this paper, we propose a novel approach named BWNH to train binary weight networks via hashing. We first transform the binary weight learning problem into a hashing problem. Then an alternating optimization algorithm is proposed to solve the hashing problem. Finally, the whole binary weight network is fine-tuned to restore accuracy. Extensive experiments on three datasets demonstrate that our algorithm outperforms state-of-art algorithms. Our main contributions are:
\begin{enumerate}[(1)]
\item We uncovered the close connection between inner-product preserving hashing and binary weight neural networks. Based on this view, training binary weight networks can be transformed into a hashing problem. To the best of our knowledge, it is the first to train binary weight CNNs via hashing. 
\item To alleviate the loss brought by hashing,  the binary codes is multiplied by a scaling factor. To solve the binary codes and scaling factor, we propose an alternating optimization method to iteratively update binary codes and scaling factor.
\item We conduct extensive experiments on CIFAR10, CIFAR100, and ImageNet. And the experimental results demonstrate that our proposed BWNH outperforms the state-of-art algorithms. 
Specifically, on ResNet-18, our BWNH achieves 3.0\% higher accuracy than the best reported binary weight networks for ImageNet classification task.
\end{enumerate}

\section{Related Work}
While deep convolutional neural networks have achieved quite good performance in many computer vision tasks, the large computational complexity and parameter size have hindered the deployment on mobile devices. A  variety of methods have been proposed to alleviate these problems.
\\
   \textbf{Pruning}  Optimal Brain Demage \cite{lecun1989optimal} and  Optimal Brain Surgeon  \cite{hassibi1993second} are early works of pruning. Both these two algorithms used Hessian matrix of loss functions, which makes them hard to scale up to large scale models. \cite{han2015learning} proposed the  deep compression framework and they introduced a three-stage pipline: pruning, trained quantization and Huffman coding. They demonstrated that such a three-stage method can reduce the parameter size of AlexNet up to 35 times. After pruning, the sparse connections of CNNs do not fit well on parallel computation. To cure this problem, a group-sparsity regularizer was proposed  \cite{lebedev2015fast}. By using the group-sparsity regularizer, they pruned the convolutional kernel tensor in a group-wise fashion. After such pruning, convolutions can be reduced to multiplications of thinned dense matrices, which can use the Basic Linear Algebra Subprograms (BLAS) to get higher speed-ups.
\\
   \textbf{Low-rank Approximation}
   Low-rank based methods assume that the featuremaps or weights of CNNs lie on a low-rank subspace. Based on this assumption, matrix or tensor decomposition methods are applied to the convolutional kernels or featuremaps \cite{denton2014exploiting,jaderberg2014speeding,denil2013predicting}. 
   By using biclustering and Singule Value Decomposition (SVD),  \cite{denton2014exploiting} achieved 1.6$\times$ speed-up of single layer. \cite{zhang2015accelerating} took nonlinear layers into consideration and used an asymmetric reconstruction method to approximate the low rank matrix.
    \cite{lebedev2014speeding} utilized CP-decomposition to approximate the 4D convolutional kernel tensor. 
    For large models such as Alexnet, their methods only processed a single layer and could not work well for the whole network. \cite{kim2015compression} proposed to use Tucker decomposition to reduce the computational complexity of CNN models. 
     By making a compromise between CP-decomposition and Tucker decomposition, Block Term Decomposition was used to accelerate the CNN models \cite{wang2016accelerating}.
\cite{novikov2015tensorizing} used the Tensor-Train format to decompose the fully connected layer, which achieved up to 7 $\times$ compression ratio of the whole network.
\\
\textbf{Quantization-based Methods}
As mentioned above, quantization-based methods can be divided into two groups: codebook-based quantization and low-bit quantization.
\cite{gong2014compressing,wu2016quantized} are typical ones of  codebook-based quantization methods. \cite{gong2014compressing} used vector quantization to compress the fully-connected layers of CNNs. And \cite{wu2016quantized} proposed an product quantization based algorithm to speed up and compress CNNs in the meantime.
For low-bit weight quantization, early works focused on using fixed-point data format to represent the weights of CNNs. \cite{gupta2015deep} introduced a stochastic rounding scheme to quantize the weights to fixed-point format. They showed that neural networks can be trained using only 16-bit fixed-point format with little degradation of classification accuracy. 
Later, a dynamic-precision data quantization method was proposed by \cite{qiu2016going}, and  8/4-bit quantization was achieved with little loss.
These methods assume that all layers share the same bit-width.  \cite{lin2015fixed} showed that it's better that different layers have different bit-width. 
Binary weight is a special case of low-bit quantization where weights are quantized into binary values.
\cite{courbariaux2015binaryconnect} proposed BinaryConnect to train CNNs with binary weights, and their method demonstrated well performance on small dataset such as MNIST, CIFAR10, and SVHN. Later, \cite{lin2015neural} proposed ternary connect to quantize the weights to ternary values, and they also quantized the back propagation. Experiments demonstrated that ternary connect achieved better result than binary connect. \cite{zhou2017incremental} proposed an incremental network quantization method which consists of three operations: weight partition, group-wise quantization and re-training. These three operations are repeated on an iterative manner, and experiments on ImageNet demonstrated that CNN models can be quantized into 5 bits without accuracy drop.
 \cite{rastegari2016xnor} proposed Binary-Weight-Networks whose weights are binarized and multiplied with a scaling factor, and they also proposed XNOR-Net by binarizing both activations and weights. \cite{cai2017deep} proposed an Halfwave Gaussian quantizer (HWGQ) for forward approximation. \cite{dong2017learning} introduced a stochastic quantization scheme which quantizing weights with a stochastic probability inversely proportional to the quantization error. 
\\
 \textbf{Hashing and Neural Networks}
  \cite{chen2015compressing} first introduced hashing methods to compress CNNs, they used the hashing trick to map the high dimensional features to a low-dimensional dictionary. The weights in the dictionary are still float-numbers, which has a large difference with our method. \cite{spring2017scalable} proposed a scalable and sustainable deep learning framework via randomized hashing. By using hash codes lookup table, they collected a small portion of neural nodes called active set. Since only these neural nodes required forward and backward propagation, computational cost was reduced. Our method is different with \cite{spring2017scalable}  in several ways. First, the hashing method is used for different goals. Our algorithm aims to learn binary weights while \cite{spring2017scalable} use hashing algorithm to select the active set. Second, our method is a learning-based (data-dependent) method while \cite{spring2017scalable} used a data-independent Locality-Sensitive Hashing (LSH) method. Third, our method is different with \cite{spring2017scalable} in the inference phase. The multiplication operation is replaced with add operation in our method while \cite{spring2017scalable} choose a subset of neural nodes to do forward propagation.
      
 \section{Our Method}
 In this section, we first introduce the inner-product preserving hashing, and uncover the close connection between inner-product preserving hashing and learning binary convolutional kernels. Then we give details about how the objective is transformed from learning binary weights to learning hashing codes. A new objective function is proposed to compensate the accuracy loss brought by hashing codes, then an alternating optimization method is introduced to solve the new objective function.
Finally, we present our whole training scheme.
 \subsection{Inner-product preserving hashing}
 Given two sets of points $\mathbf{X}\in \mathbb{R}^{S\times M}$ and $\mathbf{W}\in \mathbb{R}^{S\times N}$ where $\mathbf{X_i}\in \mathbb{R}^{S\times 1}$ and $\mathbf{W_i} \in \mathbb{R}^{S\times 1}$ represents $i^{th}$ point of $\mathbf{X} $ and $\mathbf{W} $ respectively,
 we denote the inner-product similarity of $\mathbf{X}$ and $\mathbf{W}$ as $\mathbf{S}\in \mathbb{R}^{M\times N}$.  \cite{shen2015learning} proposed the inner-product preserving hashing by solving the following objective function:
 \begin{equation}\label{hash}
 \min \,\, \lVert \mathbf{S}- h(\mathbf{X})^{\mathrm{T}}g(\mathbf{W})\rVert _F^2
 \end{equation}
 where $h(\cdot)$ and $g(\cdot)$ are hash functions for $\mathbf{X}$ and $\mathbf{W}$ respectively.
 
 \subsection{Connection between hashing and binary weights}
Suppose we have an $L$-layer pre-trained CNN model e.g. Alexnet, and $\mathbf{X} \in \mathbb{R}^{S\times M} $ is the input featuremap for $l^{th}$ layer of the CNN model. We denote the real-value weights of $l^{th}$ layer as $\mathbf{W}\in \mathbb{R}^{S\times N}$, 
and our goal is to get binary weight $\mathbf{B} \in \lbrace{-1,+1\rbrace}^{S\times N}$ for  $l^{th}$ layer of CNN model.  A naive method is to optimize the following objective function:
\begin{equation}\label{binary}
\begin{aligned}
  &\min\,\, L(\mathbf{B})= \lVert \mathbf{W}-\mathbf{B}\rVert _F^2\\
   &\,\, s.t. \quad \mathbf{B}\in\lbrace{+1,-1\rbrace}^{S\times N}
   \end{aligned}
\end{equation}
 where the solution is $\mathbf{B}= sign(\mathbf{W})$.
Directly binarizing $\mathbf{W}$ would cause severe accuracy drops, another choice is  to minimize the quantization error of inner-product similarity:    
 \begin{equation}\label{loss_func1}
 \begin{aligned}
&\min\,\, L(\mathbf{B})= \lVert \mathbf{X}^{\mathrm{T}}\mathbf{W}- \mathbf{X}^{\mathrm{T}}\mathbf{B}\rVert _F^2\\
 &\,\, s.t. \quad \mathbf{B}\in\lbrace{+1,-1\rbrace}^{S\times N}
 \end{aligned}
\end{equation}
Note Equation \eqref{loss_func1} has a close connection with Equation \eqref{hash}, let $\mathbf{S}=\mathbf{X}^{\mathrm{T}}\mathbf{W}$, $\mathbf{B}=g(\mathbf{W})$ and $h(\mathbf{X})=\mathbf{X}$, then Equation \eqref{loss_func1} is equal to Equation \eqref{hash}. In other words, training binary weight networks can be intrinsically transformed into a hashing problem. We notice that $h(\cdot)$ is an identity function, which means that we don't learn the hash codes for $\mathbf{X}$. This is commonly used for asymmetric distances calculation (ADC) in the hashing area. Now we have connected binary weight networks with hashing together,  thus we can solve binary weight $\mathbf{B}$ by borrowing methods from hashing. 

However, solving Eq. \eqref{loss_func1} still can cause somewhat accuracy drops. Inspired by  \cite{rastegari2016xnor}, we multiply a scaling factor to each hashing codes $\mathbf{B_i}$:
\begin{equation}
g(\mathbf{W})= \mathbf{B}\mathbf{A}
\end{equation}
 where $\mathbf{A}$ is a diagonal matrix and $\alpha_i=\mathbf{A_{ii}}$ is the scaling factor for $\mathbf{B_i}$. Finally, our objective function is:
 
\begin{equation}\label{final_func}
\begin{aligned}  
\min\,\, L(\mathbf{A},\mathbf{B})&= \lVert \mathbf{S}-\mathbf{X}^{\mathrm{T}}\mathbf{B}\mathbf{A}\rVert _F^2\\
&=\sum\limits_i^N \lVert \mathbf{S_i}-\alpha_i\cdot\mathbf{X}^{\mathrm{T}}\mathbf{B_i}\rVert _F^2
\end{aligned}
\end{equation}
where $\mathbf{S}=\mathbf{X}^{\mathrm{T}}\mathbf{W}$  and $\mathbf{S_i} \in \mathbb{R}^{M\times1}$ is $i^{th}$ column vector of $\mathbf{S}$.
The Eq. \eqref{final_func} can be easily divided into $N$ independent sub-problems:
\begin{equation}\label{subprob_func}
\begin{aligned}
&\min\,\, L_i(a_i,\mathbf{B_i})=\lVert \mathbf{S_i}-\alpha_i\cdot\mathbf{X}^{\mathrm{T}}\mathbf{B_i}\rVert _F^2\\
&\,\, s.t. \quad \mathbf{B_i}\in\lbrace{+1,-1\rbrace}^{S\times 1}
\end{aligned}
\end{equation}
Here we propose to use an alternating optimization method to solve Eq.\eqref{subprob_func}, i.e. update binary codes $\mathbf{B_i}$ with scaling factor $\alpha_i$ fixed, and vice versa. 
\\
\textbf{Initialization of $\mathbf{B_i}$ and $\alpha_i $}
At the beginning of alternating optimization method, we initialize $\mathbf{B_i}$ with $sign(\mathbf{W_i})$. For $\alpha_i$, we take the 
average $L1$ norm of $\mathbf{W_i}$ as initialization.
\\
\textbf{Update $\alpha_i$ with $\mathbf{B_i}$ fixed }
By expanding Eq.\eqref{subprob_func}, we have
\begin{equation}
\begin{aligned}  
\min\,L_i(\mathbf{\alpha_i})
&=const+\alpha_i^2\lVert \mathbf{X}^{\mathrm{T}}\mathbf{B_i}\rVert _F^2-2\alpha_i\mathbf{S_i}^\mathrm{T}\mathbf{X}^{\mathrm{T}}\mathbf{B_i}.
\end{aligned}
\end{equation}
Then the derivative of $L_i(\mathbf{\alpha_i},\mathbf{B_i})$ w.r.t $\alpha_i$ is:

\begin{equation} 
\frac{\partial L_i(\mathbf{\alpha_i}) }{\partial{\alpha_i}}=2\alpha_i\lVert \mathbf{X}^{\mathrm{T}}\mathbf{B_i}\rVert _F^2-2\mathbf{S_i}^\mathrm{T}\mathbf{X}^{\mathrm{T}}\mathbf{B_i}\\ 
\end{equation}
By setting it to zero, we get the solution of $\alpha_i$:
\begin{equation}\label{alpha_solution} 
\alpha_i=\frac{\mathbf{S_i}^\mathrm{T}\mathbf{X}^{\mathrm{T}}\mathbf{B_i}}{\lVert \mathbf{X}^{\mathrm{T}}\mathbf{B_i}\rVert _F^2}
\end{equation}
\textbf{Solving $\mathbf{B_i}$ with $\alpha_i$ fixed}
By expanding Equation \eqref{subprob_func}, we can get:
 \begin{equation}\label{solve_b} 
\begin{aligned}  
&\min\,\, L_i(\mathbf{B_i})
=const+ \lVert \mathbf{Z}^{\mathrm{T}}\mathbf{B_i}\rVert _F^2-2\Tr{(\mathbf{B_i}^\mathrm{T}\mathbf{q})} \\
&\,\, s.t. \quad \mathbf{B_i}\in\lbrace{+1,-1\rbrace}^{S\times 1}
\end{aligned}
\end{equation}
where $\mathbf{Z}=\alpha\cdot\mathbf{X}$ ,$\Tr()$ is the trace norm, and $\mathbf{q}=\alpha\cdot \mathbf{X}\mathbf{S_i}$. Eq. \eqref{solve_b} can be solved by \emph{discrete cyclic coordinate descent} (DCC) method which is proposed in \cite{shen2015supervised} for solving hashing codes. 
Let $b$ be the $j^{th}$ element of $\mathbf{B_i}$, and $\mathbf{B_i}^{\prime} $ the column vector of $\mathbf{B_i}$ excluding $b$. Similarly we denote the $j^{th}$ element of $\mathbf{q}$ as $\mathbf{q_j}$, and let $\mathbf{q}^{\prime}$ as the $\mathbf{q}$ excluding $\mathbf{q_j}$. Let  $\mathbf{v^\mathrm{T}}$ be the  $j^{th}$ row of matrix $\mathbf{Z}$ and $\mathbf{Z}^{\prime}$ be matrix $\mathbf{Z}$ excluding $\mathbf{v^\mathrm{T}}$. Then problem \eqref{solve_b} can be written as:
\begin{equation} 
\begin{aligned}
&\min\,\, (\mathbf{B_i}^{\prime\mathrm{T}}\mathbf{Z}^{\prime}\mathbf{v}-\mathbf{q_j})b\\
 &\,\, s.t. \quad b\in\lbrace{+1,-1\rbrace}
\end{aligned}
\end{equation}
Then we can get the solution for the $j^{th}$ element of $\mathbf{B_i}$:
\begin{equation}\label{b_solution} 
b=sign(\mathbf{q_j}-\mathbf{B_i}^{\prime\mathrm{T}}\mathbf{Z}^{\prime}\mathbf{v})
\end{equation}
By using this method, each element of $\mathbf{B_i}$ can be iteratively updated with other $S-1$ elements of $\mathbf{B_i}$ fixed. 

\begin{figure*}
\centering
\caption{The optimization loss vs Iteration using the proposed alternating optimization method on different networks}
\label{fig:loss}
\begin{subfigure}{0.35\linewidth}
\label{fig:a}
\includegraphics[width=\linewidth]{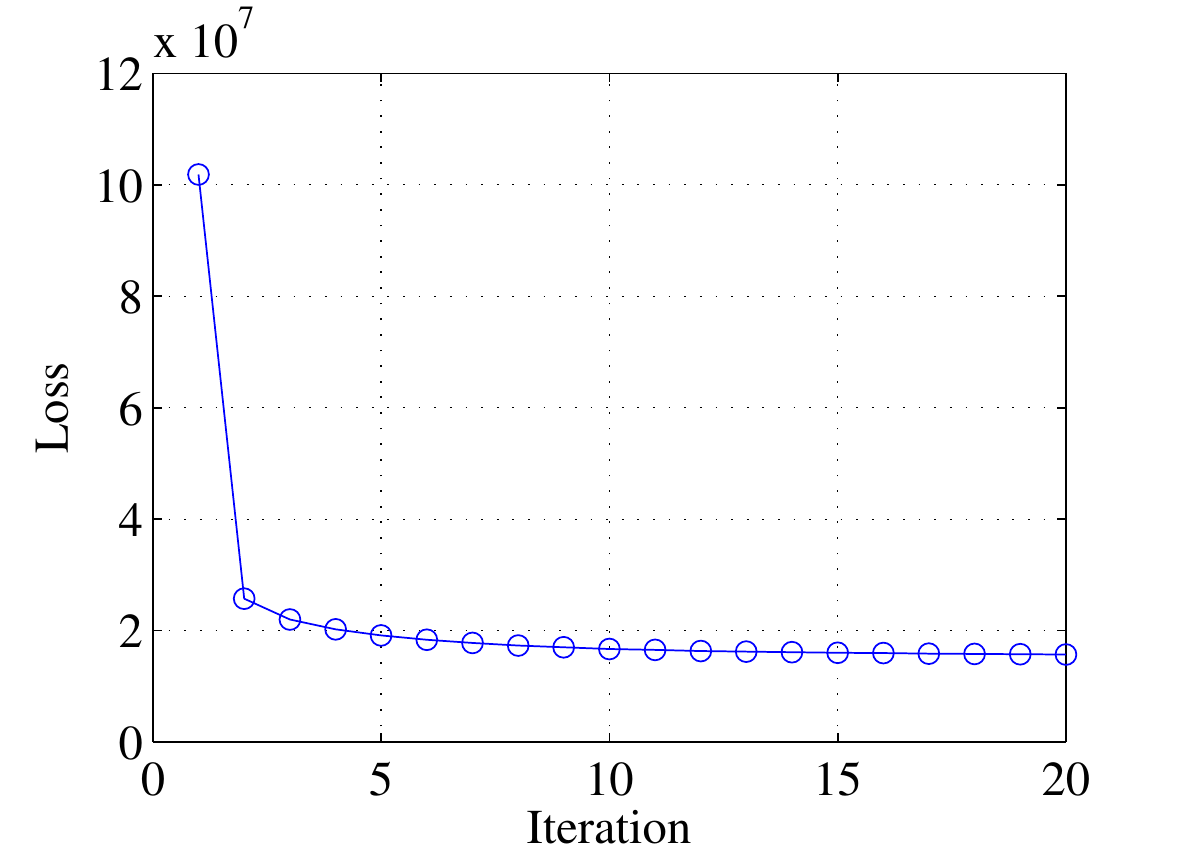}
\caption{optimization loss of VGG9 on CIFAR10    }
\end{subfigure}
\hspace{-0.04\linewidth}
\begin{subfigure}{0.35\linewidth}
\label{fig:b}
\includegraphics[width=\linewidth]{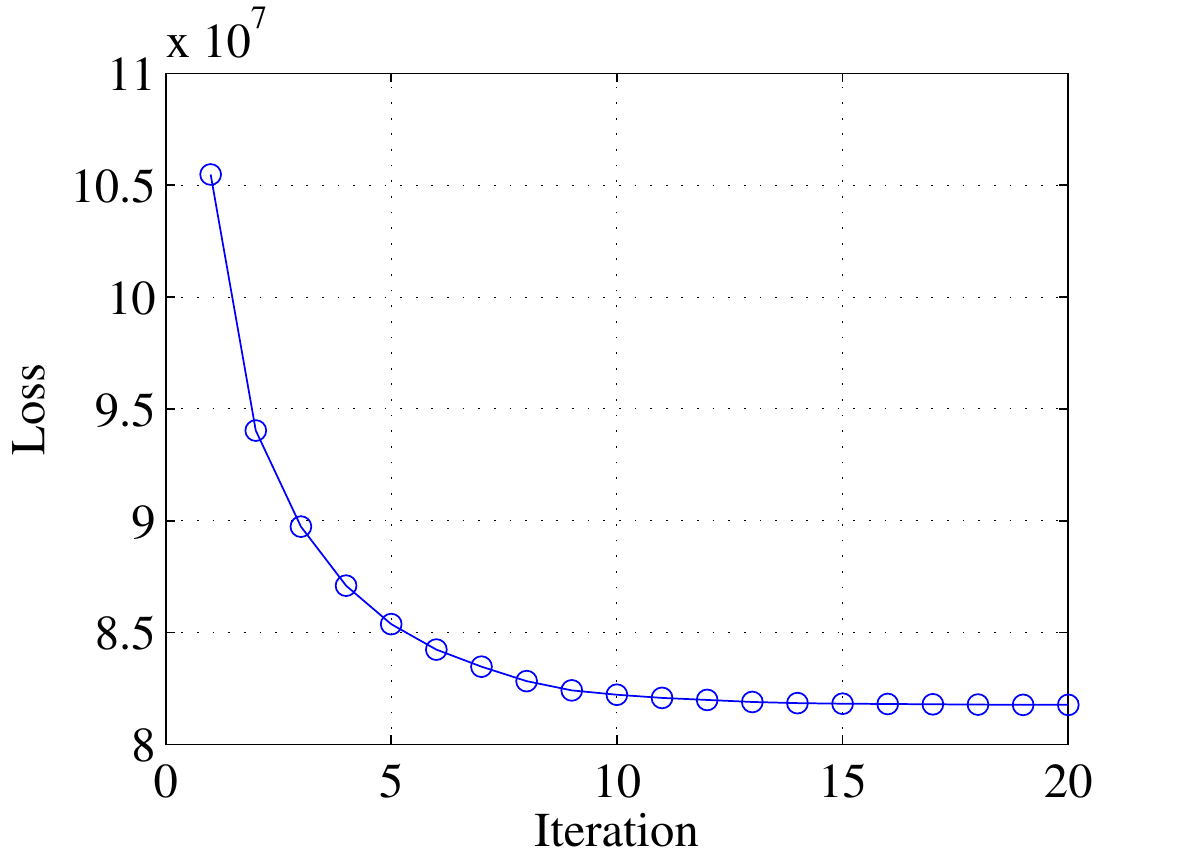}
\caption{optimization loss of AlexNet }
\end{subfigure}
\hspace{-0.04\linewidth}
\begin{subfigure}{0.35\linewidth}
\label{fig:c}
\includegraphics[width=\linewidth]{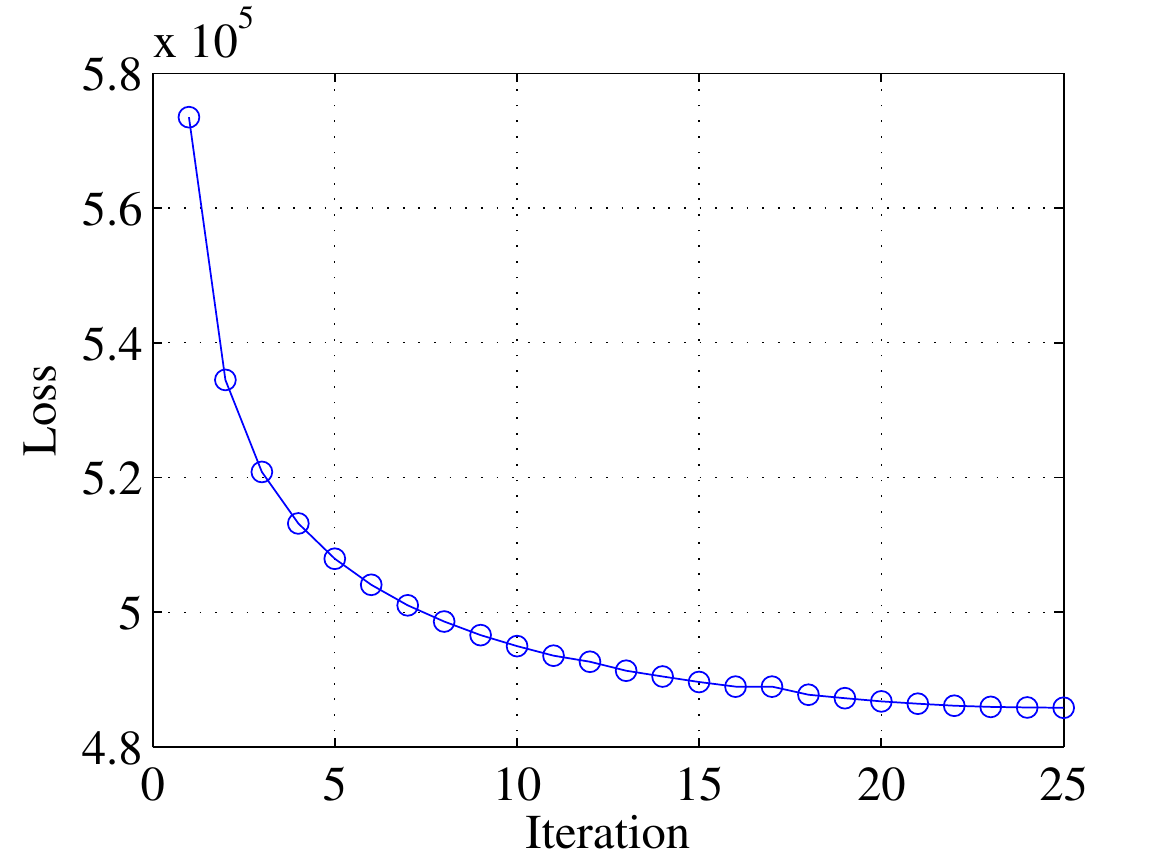}
\caption{optimization loss of ResNet-18 }
\end{subfigure}
 \end{figure*}

  The convergence of our proposed alternating optimization method is guaranteed theoretically. And It can be easily proven 
since every update step decreases the objective function value and the objective function has a lower bound. Empirical results demonstrate that the algorithm takes a few iterations to converge. Figure \ref{fig:loss} shows the convergence curves on different convolutional neural networks via the proposed alternating optimization method.
It's clear that our algorithm get converged in a few iterations. 
\subsection{Layer-wise optimization} 
By using the proposed alternating optimization method, we optimize the binary weight layer by layer. One concern is the quantization error will be accumulated across multiple layers. More specifically, quantizing the weights of $l^{th}$ layer will cause quantization error of output featuremaps which are the input featuremaps of $(l+1)^{th}$ layer, consequently affects the optimization procedure of $(l+1)^{th}$ layer. To solve this problem, the hashing codes are supposed to adapt to the input featuremaps which are affected by binary weights of the previous layer.

Here we adopt the similar training scheme as \cite{wu2016quantized}. Suppose we have a pre-trained $L$-layer CNN model and a binarized CNN model whose first $l^{th}$ layers have been binarized, we denote the input featuremaps of $(l+1)^{th}$ layer for pre-trained CNN model and binarized CNN model as $\mathbf{X}^{l+1}$ and  $\tilde{\mathbf{X}}^{l+1}$. The objective function would be:
\begin{equation}
\begin{aligned}  
\min L(\mathbf{A},\mathbf{B})&= \lVert (\mathbf{X}^{l+1})^\mathrm{T}\mathbf{W}^{l+1}-(\tilde{\mathbf{X}}^{l+1})^{\mathrm{T}}\mathbf{B}^{l+1}\mathbf{A}^{l+1}\rVert _F^2\\
&=\lVert \mathbf{S}^{l+1}-(\tilde{\mathbf{X}}^{l+1})^{\mathrm{T}}\mathbf{B}^{l+1}\mathbf{A}^{l+1}\rVert _F^2\\
\end{aligned}
\end{equation}
On one hand, the target similarity matrix $\mathbf{S}^{l+1}$( in a hashing view) is calculated between $\mathbf{W}^{l+1}$ and input featuremaps $\mathbf{X}^{l+1}$. On the other hand, the realistic similarity is calculated between  $\tilde{\mathbf{X}}^{l+1}$ and binary codes $\mathbf{B}^{l+1}$. Thus the binary codes is trained to adapt to quantization error of the input featuremaps. By such a layer by layer  training scheme, the quantization error explosion is avoided. 

\subsection{The Whole Training Scheme}
For a given pre-trained CNN model, we first use the proposed method to binarize weights of CNN model layer by layer. Then we fine-tune the binarized CNN model to get a better result. For the fine-tuning procedure, the weights of convolution layer is initialized by the learned binary codes. The scaling factor is used to initialize the weights of scale layer which is added right after the convolutional layer. We summarize the overall training algorithm in Algorithm  \ref{algorithm1}.
\begin{algorithm}
\caption{Training Binary weight Convolutional Neural Networks via Hashing } 
\label{algorithm1}
\KwIn{Pre-trained convolutional neural networks weights $\lbrace{\mathbf{W}^{l}\rbrace}_{l=1}^{L}$ and Max\_Iter} 
\KwOut{Learned binary weights $\lbrace{\mathbf{B}^{l}\rbrace}_{l=1}^{L}$ and scaling factors $\lbrace{\mathbf{A}^{l}\rbrace}_{l=1}^{L} $}

\For{$l=1;l \le L$} 
{
	Sampling a mini-batch images from database \\
	Forward propagation to get $\tilde{\mathbf{X}}^{l}$ and $\mathbf{X}^{l}$\\
	
	Calculate $S$ with  $\tilde{\mathbf{X}}^{l}$\\
	\For{$i=1;i \le N$} 
{
	Initialize $\mathbf{B_i}$ with $sign(\mathbf{W_i})$ \\
	Initialize $\alpha_i$ with  mean $L1$ norm of $\mathbf{W_i}$\\
	\While{iter $\le$ Max\_Iters} 
	{ 
		Update $\alpha_i$ with Eq.\eqref{alpha_solution} \\
		\For{$j=1;j \le S$} 
		{
			Update $j^{th}$ element of $\mathbf{B_i}$ with Eq.\eqref{b_solution} \\
	    }
	 }
	 }
 }
 \For{$l=1;l \le L$} 
{
Initialize  $l^{th}$ layer of binarized CNN model with $\mathbf{B}^{l}$\\
 Add a scale layer right after the $l^{th}$ layer \\
 Initialize weights of the scale layer with $\mathbf{A}^{l}$\\
}
Fine-tune the binarized CNN model\\
return   $\lbrace{\mathbf{B}^{l}\rbrace}_{l=1}^{L}$ and $\lbrace{\mathbf{A}^{l}\rbrace}_{l=1}^{L}$; 
\end{algorithm}

\section{Experiments}
In this section, we first give details of experiment settings including datasets, network architectures, training settings and so on.
Then experimental results on three datasets are analysed, showing that our proposed method outperforms the state-of-art algorithms. Finally, we analyse the effect of scaling factor $\mathbf{A}$.
\subsection{Datasets}  
To evaluate our proposed method, we conduct extensive experiments on three public benchmark datasets including CIFAR10, CIFAR100, and ImageNet.
\begin{itemize}
\item \textbf{CIFAR10} dataset consists of 60,000 colour images in 10 classes. Each class contains  6000 images in size 32$\times$ 32. There are 5000 training images and 1000 testing images per class.
\item \textbf{CIFAR100} dataset is like CIFAR10 except it has 60,000 colour images in 100 classes. There are 500 training images and 100 testing images per class.
\item \textbf{ImageNet} dataset (ILSVRC2012) has about 1.2M training images from 1000 classes and 50,000 validation images. Compared to CIFAR10 and CIFAR100, the images in ImageNet have higher resolution and complex background.
\end{itemize}
\subsection{Network Architectures}
 For ImageNet (ILSVRC2012), two state-of-art networks i.e. AlexNet and ResNet-18 are adopted to evaluate the proposed method. For CIFAR10 and CIFAR100, we adopt the VGG-9 network following \cite{dong2017learning}. \\
\textbf{AlexNet} 
AlexNet consists of 5 convolutional layers and three fully-connected layers. Following \cite{rastegari2016xnor,dong2017learning}, we use AlexNet coupled with batch normalization layers.
\\
\textbf{ResNet-18} Recently \cite{he2015deep} proposed ResNet architecture which is more efficient and powerful. Base on this architecture, very deep convolutional neural networks can be trained efficiently. The ResNet architecture is built with residual blocks, which is quite different with AlexNet. Here we adopt the ResNet-18 architecture. \\
\textbf{VGG-9}  
 The architecture of VGG-9 is denoted as ``(2$\times$64C3)-MP2-(2$\times$128C3)-MP2-(2$\times$256C3)-MP2-(2$\times$512C3)-10FC-Softmax". Here `64C3' denotes convolutional layer with 64 kernels of size 3$\times$3. MP denotes max-pooling layer and FC denotes the fully-connected layer.
Batch Normalization layer is added after each convolutional or fully-connected layer. 
 \subsection{Training Settings}
 We implement our proposed method based on the Caffe framework, and the proposed alternating optimization algorithm is implemented using CUDA. All experiments are conducted on a GPU Server which has 8 Nvidia Titan Xp GPUs.\\ 
During layer-wise optimization, we set maximum iterations of the proposed alternating optimization method to 20 which is enough for training according to Figure \ref{fig:loss}.
 We adopt different fine-tuning settings for different network architecture.
 \\
 \textbf{AlexNet} We fine-tune AlexNet using a SGD solver with momentum=0.9, weight decay=0.0005. The learning rate starts at 0.001, and is divided by 10  after 100k, 150k, and 180k iterations. The network is fine-tuned for 200k iterations with batch-size equals to 256. Before training, images are resized to have 256 pixels at their smaller side. Random cropping and mirroring are adopted in the training stage and center cropping is used in the testing stage.  
\\
 \textbf{ResNet-18} We fine-tune the ResNet-18 using a SGD solver with momentum=0.9, weight decay=0.0005. The learning rate starts at 0.0005, and is divided by 10 every 200k iterations. We run the training algorithm for 650k iterations with batch size equal to 128. We use random cropping and mirroring for data augmentation. Like AlexNet, images are resized to have 256 pixels at their smaller side.
 \\
 \textbf{VGG-9}  We use a SGD solver with momentum=0.9, weight decay=0.0001 for fine-tuning the VGG-9 network. The learning rate starts at 0.1, and is divided by 10  every 15K iterations. Following \cite{dong2017learning}, the network is fine-tuned for 100K iterations with batch-size equals to 100. 
   \begin{table} [!ht]
\centering
\caption{Test error rate of VGG9 on CIFAR10 and CIFAR100}
\begin{tabular}{c|c|c}
\hline

\multirow{2}{*}{Method}   & \multicolumn{2}{c}{Test error rate}\\ \cline{2-3}
							& CIFAR10 & CIFAR100 \\ \hline \hline
Full-Precision &  9.01&   30.45 \\	\cline{1-3} 	\hline  			
BinaryConnect &  11.15&    37.70 \\
\hline  
BWN&   10.67& 37.68       \\ \cline{1-3} 
\hline  
SQ-BWN&   9.40& 35.25      \\ \cline{1-3}  
\hline \hline
BWNH (Ours)&   \textbf{9.21}& \textbf{34.35}       \\ \cline{1-3}  	 
\end{tabular}
\label{tab:vgg9}
\end{table}
 \subsection{Experimental Results}
  To evaluate our proposed method, we compare our method with BC \cite{courbariaux2015binaryconnect}, BWN \cite{rastegari2016xnor}, SQ-BWN \cite{dong2017learning},  and HWGQ-BWN \cite{cai2017deep}. Table. \ref{tab:vgg9} shows the classification accuracy of VGG9 network on CIFAR10 and CIFAR100 dataset via different methods. From Table. \ref{tab:vgg9}, it's clear that our proposed method outperforms other state-of-art algorithms.
  
  Since CIFAR10 and CIFAR100 both are small dataset, we mainly verify our proposed method and tune the hyper-parameters  on these two datasets. Here we focus on the experiment results on ImageNet.
  Table. \ref{tab:alexnet} demonstrates the Top1 and Top5 classification accuracy of AlexNet on ImageNet dataset for different methods. And our proposed method outperforms the state-of-art methods in both Top1 and Top5 accuracy.
  
   Table. \ref{tab:resnet} compares our proposed BWNH with other methods on ResNet-18 network. We can find that our proposed method outperforms the state-of-art method by a large margin (3\% in top1 accuracy).
\begin{figure}
\centering
   \includegraphics[width=1\linewidth]{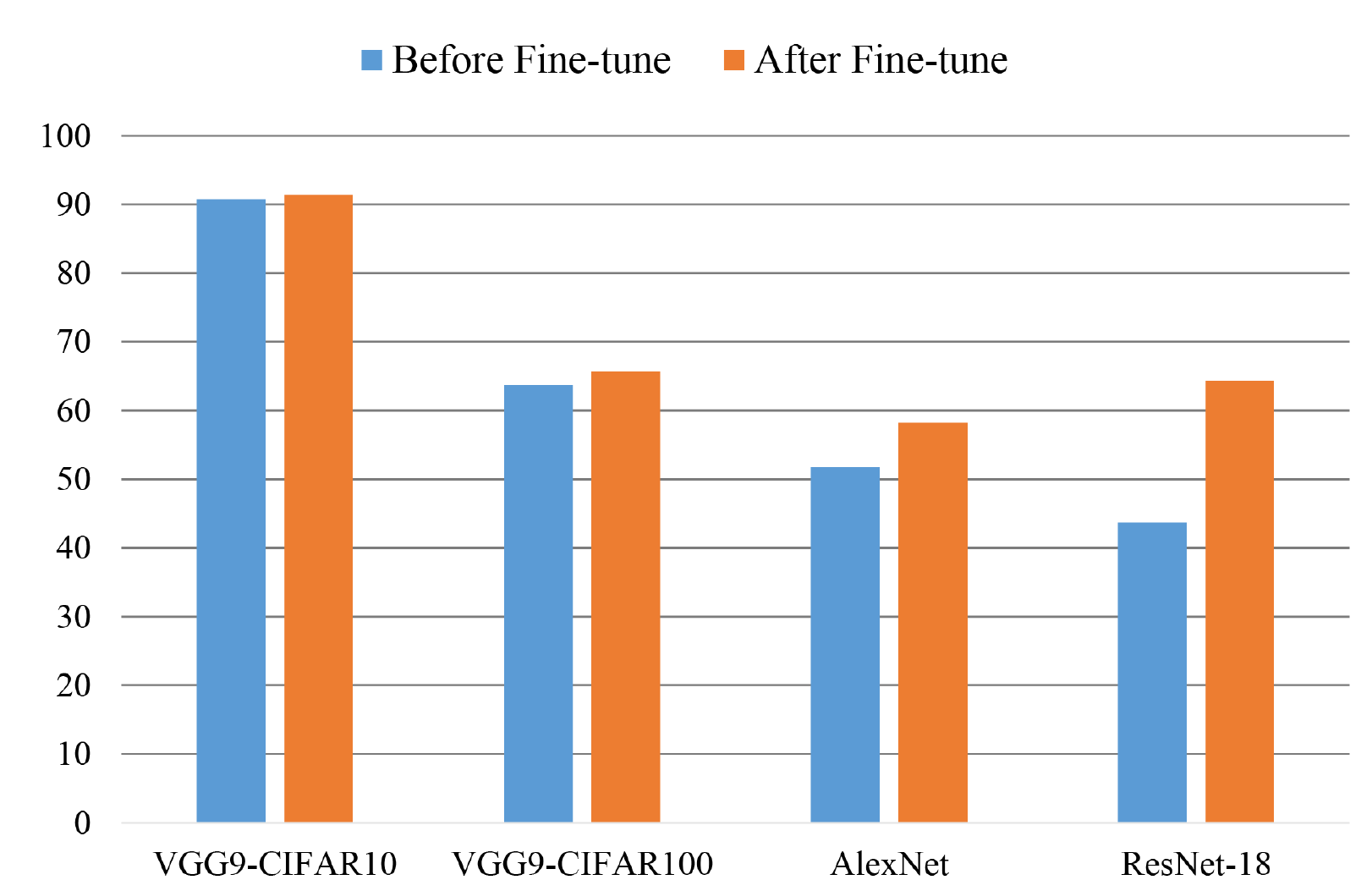}
   \caption{Top1 Accuracy of VGG9, AlexNet, and ResNet-18 with or without fine-tuning }
\label{finetune}
\end{figure}
\begin{table}[!ht]
\centering
\caption{Classification Accuracy of AlexNet for different methods}
\begin{tabular}{c|c|c} 
\hline
\multirow{2}{*}{Method}   & \multicolumn{2}{c}{Classification Accuracy}\\ \cline{2-3}
							& Top1 & Top5 \\ \hline \hline
							
BinaryConnect &  35.4&    61.0 \\
\hline  
BWN&   56.8& 79.4       \\ \cline{1-3} 
\hline  
SQ-BWN&   51.2& 75.1       \\ \cline{1-3} 
\hline  
HWGQ-BWN & 52.4& 75.9  \\ \cline{1-3} 
\hline \hline
BWNH (Ours)&   \textbf{58.5}& \textbf{80.9}       \\ \cline{1-3}  
\end{tabular}
\label{tab:alexnet}
\end{table}
\begin{table} [!ht]
\centering
\caption{Classification Accuracy of ResNet-18 for different methods}
\begin{tabular}{c|c|c}
\hline
\multirow{2}{*}{Method}   & \multicolumn{2}{c}{Classification Accuracy}\\ \cline{2-3}
							& Top1 & Top5 \\ \hline \hline							
Full-Precision &  69.3&    89.2 \\
\hline  
BWN&   60.8& 83.0       \\ \cline{1-3} 
\hline  
SQ-BWN&   58.3& 81.6       \\ \cline{1-3}  
\hline  
HWGQ-BWN & 61.3& 83.9  \\ \cline{1-3} 
\hline \hline
BWNH (Ours)&   \textbf{64.3}& \textbf{85.9}       \\ \cline{1-3}  			 		
\end{tabular}
\label{tab:resnet}
\end{table}

Figure. \ref{finetune} shows the Top1 accuracy of different networks by using proposed BWNH with or without using a fine-tuning step. From Figure. \ref{finetune}, we notice that our proposed BWNH has achieved a relatively high accuracy without the fine-tuning step, which demonstrates the usefulness of learning binary weights via hashing. The binary weights learned by hashing can be used as initialization of CNNs, and a fine-tuning step will improve the accuracy of networks. Other training binary weights algorithms can also be combined with our methods by simply using our learned binary weights as initialization, which can get higher accuracy.
\begin{figure}
\centering
   \includegraphics[width=1\linewidth]{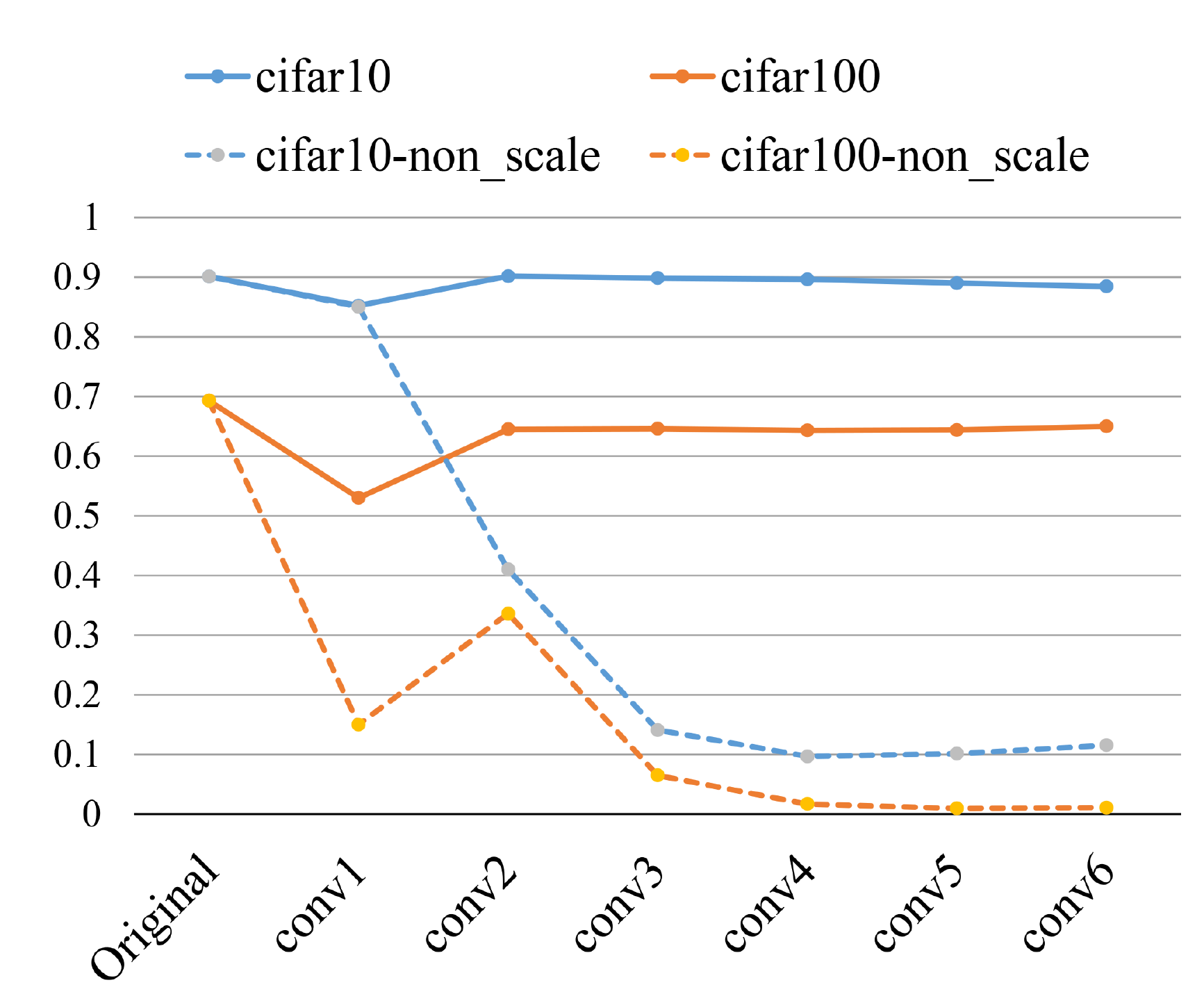}
   \caption{Accuracy of VGG9 Network on CIFAR10 and CIFAR100 after layer-wise optimization from \emph{conv1} to \emph{conv6}}
\label{scale_factor}
\end{figure}
%
\subsection{The Effect of Scaling Factor}
In this subsection, we explore the effect of scaling factor.
Figure. \ref{scale_factor} shows the accuracy of VGG9 network on CIFAR10 and CIFAR100  by using proposed BWNH  with or without scaling factor. We denote the result without using scaling factor as \emph{non\_scale}. From Figure. \ref{scale_factor}, it's clear that the scaling factor is very important for the proposed method. Without the scaling factor, the accuracy of network degrades very quickly and reaches the \emph{random} accuracy (~0.1 for CIFAR10 and ~0.01 for CIFAR100) after optimizing several layers. Besides, the scaling factor can be merged into the Batch Normalization layer in the inference phase, thus it won't add extra memory or storage overhead. Another interesting phenomenon in Figure. \ref{scale_factor} is that the accuracy after optimizing \emph{conv1} and \emph{conv2} is higher than the accuracy after optimizing \emph{conv1}. This is because the binary weights in \emph{conv2} compensates the accuracy drop by adapting to the input featuremaps generated by binary weights in \emph{conv1}.
 
\section{Conclusion and Future Work}
 In this paper, we first uncovered the close connection between inner-product preserving hashing and binary weight networks and showed that training binary weight networks can be transformed into a hashing problem. A scaling factor is multiplied to the binary codes to improve the accuracy, then we propose an alternating optimization method to solve the problem. Experiments on CIFAR10, CIFAR100, and ImageNet show that our proposed method outperforms the state-of-art algorithms. 
 
 At present, our method aims to quantize the weights of CNNs to 1 bit (-1 or +1). In fact, our method can also be used to train a binary neural network (BNN) where the featuremaps and weights of CNN are all quantized to 1 bit. 
  In the future, we will try to train binary neural networks using our proposed method. 
\section{Acknowledgements}
This work was supported in part by National Natural Science Foundation of China (No.61332016), the Scientific Research Key Program of Beijing Municipal Commission of Education (KZ201610005012), Jiangsu Key Laboratory of Big Data Analysis Technology, and Hubei Key Laboratory of Transportation Internet of Things.
\bibliography{AAAI2018_Hu} 
\bibliographystyle{aaai}
\end{document}